\documentclass{article}
\usepackage{spconf,amsmath,graphicx,amssymb}
\usepackage{tabularx}
\usepackage[numbers,sort&compress]{natbib}

\title{Robust and computationally-efficient anomaly detection USING POWERS-of-TWO Networks}
\name{
  \parbox{110mm}{
    \centering Usama Muneeb$^{\star}$, Erdem Koyuncu$^{\star}$, Yasaman Keshtkarjahromi$^{\dagger}$, Hulya Seferoglu$^{\star}$, Mehmet Fatih Erden$^{\dagger}$, A. Enis Cetin$^{\star}$ \thanks{This research is funded by Seagate US LLC.}
  }
}
\address{$^{\star}$ Dept. of Electrical and Computer Engineering, University of Illinois at Chicago\\ Chicago, IL 60607\\
 $^{\dagger}$ Seagate Technology LLC\\Shakopee, MN 55379}
\begin{document}
\ninept
\maketitle
\begin{abstract}
Robust and computationally efficient anomaly detection in videos is a problem in video surveillance systems.
We propose a technique to increase robustness and reduce computational complexity in a Convolutional Neural Network (CNN) based anomaly detector that utilizes the optical flow information of video data.
We reduce the complexity of the network by denoising the intermediate layer outputs of the CNN and by using powers-of-two weights, which replaces the computationally expensive multiplication operations with bit-shift operations. Denoising operation during inference forces small valued intermediate layer outputs to zero. The number of zeros in the network significantly increases as a result of denoising, we can implement the CNN about 10\% faster than a comparable network while detecting all the anomalies in the testing set. It turns out that denoising operation  also provides robustness because the contribution of small intermediate values to the final result is negligible.
During training we also generate motion vector images by a Generative Adversarial Network (GAN) to improve the robustness of the overall system.  We experimentally observe that the resulting system is robust to background motion.


\end{abstract}
\section{Introduction}
\label{sec:intro}
\vspace{-0.1in}
Generative Adversarial Networks (GANs) \cite{DBLP:conf/nips/GoodfellowPMXWOCB14} are networks trained via an adversarial process primarily to capture the data distribution and to generate artificial data or objects that look like real objects.
In recent years GANs have also been used to perform anomaly detection \cite{DBLP:journals/corr/abs-1802-06222} \cite{DBLP:conf/icip/RavanbakhshNSMR17}.

Owing to the huge computational complexity of GAN based anomaly detection techniques, this paper aims to explore the potential of using a powers-of-two \cite{DBLP:conf/dac/TannHBR17} \cite{DBLP:conf/dac/DingLCMB19} Convolutional Neural Network (CNN) in anomaly detection with no computational overhead during inference. In this paper, we use the GAN structure during only training to generate motion vector images for the computationally efficient Convolutional Neural Network (CNN) with power-of-two coefficients. Because of the special structure of the filter coefficients the convolutional filters can be implemented using only bit-shift operations without performing any multiplications.

Our CNN structure is more efficient than other power-of-two networks and GAN-improved CNNs because (i) we not only replace multiplications with bit-shift operations but also (ii) prune the intermediate outputs of the network layers by denoising. As the number of zeros significantly increase as a result of denoising, we can implement the CNN about 30\% faster than a comparable power-of-two CNN during inference. While the primary motive is to reduce computational complexity, it also achieves robustness against
background motion due to wind, leaves etc. This is
because contributions of small valued intermediate output values to the final result are eliminated by denoising. It should be noted that denoising the network layers is different from filter weight pruning which is often used in deep neural networks. Instead of the weights, we prune the outputs of the intermediate layers in this paper.

We project the intermediate output layer vectors to $\ell_1$-balls to achieve denoising. This is equivalent to soft-thresholding, which does not add any significant computational load. Motion vectors due to background objects are very small and should not contribute to the anomaly detection process. Intermediate layer denoising removes the effects of background objects and their motion to the final result. This approach is similar to the wavelet denoising \cite{DBLP:journals/tit/KrimTMD99} in which band-pass and high-pass filtered data is soft-thresholded in the wavelet domain. In this paper, we soft-threshold the intermediate layer outputs which are obtained after convolutional filtering and nonlinear activation.

The paper is organized as follows. In Section 2, we describe the motion vector based anomaly detection network. In Section 3, we describe the $\ell_1$-ball based denoising that we use to denoise the intermediate layer outputs of the CNN. In Section 4, we present the simulation results. In Section 5, we conclude the paper.

\vspace{-0.1in}
\section{Motion Vector Based Anomaly Detection}
\label{sec:s2motionvectanom}
\vspace{-0.1in}
Most anomaly detection methods are based on motion information which use hand-crafted features to model normal-activity patterns
\cite{DBLP:conf/wacv/MousaviMPCM15, DBLP:conf/cvpr/MehranOS09, DBLP:conf/cvpr/MahadevanLBV10, DBLP:conf/cvpr/CongYL11, DBLP:conf/cvpr/KimG09, DBLP:books/sp/13/RaghavendraCBSM13, DBLP:conf/avss/RabieeHMKNM16, DBLP:conf/cvpr/SaligramaC12, DBLP:journals/mlc/RabieeMNR18, DBLP:journals/itiis/HuangWSFK16}.
On the other hand our method uses the entire set of motion vectors obtained from the video as in \cite{DBLP:conf/wacv/RavanbakhshNMSS18}, \cite{DBLP:journals/cviu/SabokrouFFMK18}, \cite{DBLP:journals/cviu/XuYRS17}. We also found the optical flow domain as a reasonable representation for training and evaluating our model. For example, suddenly stopping objects or fast moving objects have larger motion vectors than objects performing usual activities.
\begin{figure}[htb]
\includegraphics[width=200px]{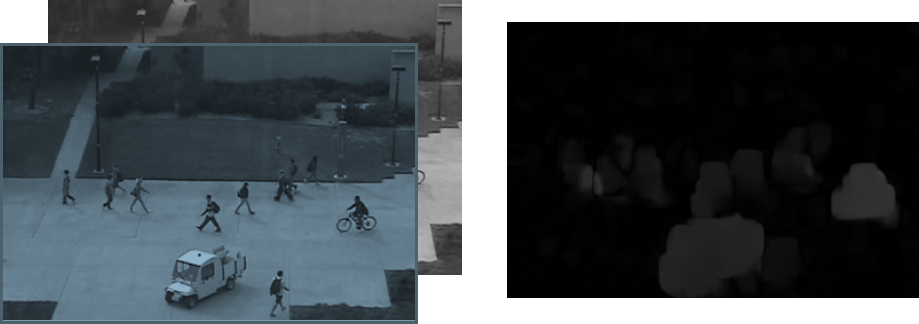}
\caption{An image from the UCSD Pedestrian database (left) and its Farneback optical flow (magnitudes) computed using the neighboring image frame.}
\label{fig:farneback}
\vspace{-0.1in}
\end{figure}
In Figure \ref{fig:farneback}, an image (left) and the corresponding optical flow image (right) from UCSD Pedestrians database \cite{DSET:UCSD/PEDS} are shown. We use the Farneback optical flow method \cite{DBLP:conf/scia/Farneback03} to find the optical flow. The optical flow image on the right shows only the magnitude information of motion vectors. Vehicle and the biker are fast moving objects in the scene (shown on the left). As a result, we observe large (and brighter) blobs due to the vehicle and the biker, which are considered to be anomalous objects in the scene. On the other hand pedestrians have relatively smaller motion vectors which translate to darker blobs on the optical flow image.
Magnitudes of motion vectors are larger for trucks, bicycles and skateboarders compared to relatively slow moving pedestrian subjects which have a smaller and not as bright representation.

Background objects (e.g. leaves, grass etc) and camera swaying due to wind add noise to the optical flow of the video. One of our motives is to make the detector resilient to such noise. The challenge is the lack of noisy training data. To this end, we use a GAN to generate the data to compensate for the lack of noisy data.

\begin{figure}
 \vspace{-0.1in}
  \centering
  \begin{tabular}{@{}c@{}}
    \includegraphics[width=.4\linewidth,height=60pt]{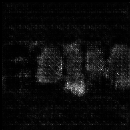} \\[\abovecaptionskip]
  \end{tabular}
  \caption{GAN generated optical flow image (magnitude)}\label{fig:realvsgan}
  \vspace{-0.1in}
\end{figure}

To improve the accuracy of the CNN we can train a GAN on optical flows and run it to generate blurry optical flow images to further train the CNN. In Figure \ref{fig:realvsgan} a GAN generated optical flow image is shown. Our hypothesis is that further training the CNN on the blurry GAN generated data will make it resilient to noise, hence making our system resilient to real life disturbances (i.e., camera swaying and background object movements).
We assume we do not know the actual noise model of the optical flow images and will not use any noisy optical flow images in training. Hence the GAN turns out to be a valuable player in giving us the ``noisy'' data for training.
 \vspace{-0.1in}
\section{Denoising the Intermediate Output Layers During Inference}
\label{sssec:s3denoising}
 \vspace{-0.1in}
 $\ell_1$ norm was used in training both GANs and CNNs in the past. For example,
it is observed that it is beneficial to combine the GAN objective function with a more traditional loss, such as the $\ell_1$ distance \cite{DBLP:conf/cvpr/PathakKDDE16} \cite{DBLP:conf/cvpr/IsolaZZE17}.
In pix2pix system (\cite{DBLP:conf/cvpr/IsolaZZE17} \cite{DBLP:journals/corr/abs-1802-06222}), the linear combination of the GAN cost function with the $\ell_1$ norm based cost function is used:
\begin{equation}
C = L_{cGAN}(G,D)+ \lambda E[||y \ - \ G(x,z)||_1]
\end{equation}
where $y$ is the output image, $x$ is the observed image and $z$ is a noise vector; $G$ and $D$ represent the generator, and the discriminator networks, respectively, $\lambda$ is the mixture parameter and
\begin{equation}
L_{cGAN}(G,D) = \mathbb{E} [\log D(x, y) + \log(1 - D(x, G(x, z))]
\end{equation}
$\ell_1$ based cost functions make the resulting solution sharper and sparser than Euclidian objective functions \cite{RUDIN1992259, 10.2307/2346178, DBLP:journals/spm/CetinT15, DBLP:conf/icip/TofighiKC14, DBLP:conf/icml/DuchiSSC08}.

\begin{figure}[htb]
\centering
\includegraphics[width=100px]{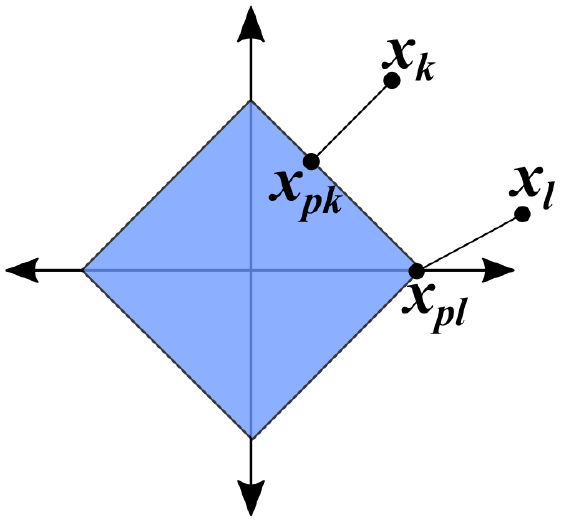}
\caption{Geometric illustration of projections onto an $\ell_1$-ball}
\label{fig:l1ball}
\end{figure}

Our approach is different from the previous work \cite{DBLP:conf/icassp/AfrasiyabiBNYYC18} in the sense that we project convolutional layer outputs to $\ell_1$-balls (Figure  \ref{fig:l1ball}) during inference. The projection "denoises" the layer outputs by eliminating small valued coefficients. This approach not only leads to a computationally efficient system by eliminating a significant portion (around 30\%) of small valued output values but also leads to a robust system because the contribution of small valued intermediate values due to background motion to the final result will be reduced.
\vspace{-0.1in}
Let $\bf x_k$ be the vector obtained after the k-th convolutional layer (filtering) and nonlinearity (activation function). We project $\bf x_k$ onto an $\ell_1$-ball of size $\epsilon$
\begin{equation}
S_1 = \{ {\bf x \mid  \left\lVert x \right\rVert _1 }\le \epsilon \}
\end{equation}
and obtain
\vspace{-0.1in}
\begin{equation}
{\bf x_{pk} = P_{S_1} (x_k) }
\end{equation}
where $P_{S_1}$ represent the orthogonal projection operator onto the $\ell_1$-ball.
 The size of the ball, $\epsilon$, may vary depending on the layer and the individual convolutional filter. Projection onto the $\ell_1$-ball can be implemented in a computationally efficient manner \cite{DBLP:conf/icml/DuchiSSC08}. It is also possible to estimate the size of the $\ell_1$-ball in an adaptive manner using the epigraph set of the $\ell_1$-ball \cite{DBLP:conf/icip/TofighiKC14} \cite{DBLP:journals/tsp/KopsinisST11} but this may increase the computational cost. We can estimate $\epsilon$ during training for each layer or individual convolutional filter.

Projection onto the $\ell_1$-ball is essentially equivalent to soft thresholding \cite{10.2307/2291512}, \cite{DBLP:journals/tit/KrimTMD99} because the last step of the projection algorithm \cite{DBLP:conf/icml/DuchiSSC08} is the soft thresholding operation:
\begin{equation}
w_{i,k} = \max \{|x_{i,k} | - \theta , 0 \}
\label{softTH1}
\vspace{-0.1in}
\end{equation}
with
\begin{equation}
\vspace{-0.05in}
u_{i,k} = \text{sgn} \{x_{i,k}\} w_{i,k}
\label{softTH2}
\end{equation}
where $x_{i,k}$ and $u_{i,k}$ are the i-th entries of  vectors $\bf x_k$ and $\bf x_{pk}$, respectively, and $\theta$ is a parameter related with the size of the $\ell_1$-ball. In Fig. 3 the projection operation onto the $\ell_1$-ball is illustrated in 2D. Parameter $\theta$ is subtracted from all the components of $\bf x_k$ to obtain $\bf x_{pk}$. In the second case, the first component of $\bf x_l$ becomes 0 after the projection.
In Donoho and Johnstone's wavelet denoising algorithm the input vector goes through a wavelet filter and it is soft-thresholded or hard-thresholded. Therefore, our idea of projecting the convolutional layer outputs is similar to adaptive wavelet denoising in the sense that the input vector goes through the filter of the neural network and soft-thresholded. In deep neural networks we have an additional nonlinearity such as RELU or Leaky-ReLU which also provides some denoising but we observed that this is not sufficient for zeroing out small valued convolutional filter outputs.
This may be due to a problem in training deep neural networks because the cost function of deep neural networks or GANs have many local optima. Soft-thresholding leads to about 30\% computational savings during inference without loosing any recognition accuracy after the nonlinear Leaky-ReLU operations. As pointed above,
we estimate the parameter $\theta$ during the training  of the neural network.
Obviously, if the nonlinearity is the RELU, then we do not need Eq.~\ref{softTH2} because all the elements are non-negative.

\section{Experimental Results}
\vspace{-0.1in}
We trained both a Deep Convolutional Generative Adversarial Network (DCGAN \cite{DBLP:journals/corr/RadfordMC15}) and a powers-of-two (Pow2) CNN using motion vectors of the UCSD Pedestrians datasets.
The training process is as described below:

{\em GAN Training}:
\label{ssec:traingan}
The topology of a DCGAN is summarized as follows:\\
- Strided convolutions are performed in discriminator network and fractional strided convolutions are carried out in the generator network of the GAN.\\
- Discriminator: LeakyReLU activations in all layers.\\
- Generator: ReLU activation for all layers except the last one, which uses Tanh function.\\
- Batch Normalization in some layers in both generator $G$ and discriminator $D$.\\
- Unlike the original DCGAN, we also add two fully connected hidden layers to the end of the discriminator.\\
The GAN is trained on the optical flows computed from the entire training dataset. UCSD Pedestrians datasets (both 1 and 2) only have normal data in the training sets.
Once trained, our GAN has learnt to generate optical flow images as shown in Figure \ref{fig:realvsgan}.
After the training, we only use the generator block to improve CNN training.

The  CNN has the same topology as the discriminator block of the DCGAN. To obtain Pow2 filter coefficients we quantize the convolutional filter coefficients to powers-of-two during each training epoch.
The CNN is also trained using the augmented data. Since we do not have any anomalous training motion images and CNN requires supervised training, we generate anomalous motion vector images artificially. This is done by creating an artificial optical flow images with a random bright blob denoting anomaly, and then feeding it to the intermediate layers of the generator to obtain a set of optical flow images. Additionally, We use a mini-batch size of 64 and train until the Discriminator Loss ($L_D$) is below 0.01. Training is stopped soon after the loss falls below the desired value to avoid overfitting. This model is saved for evaluation as a baseline
\cite{DBLP:journals/corr/MolchanovTKAK16}.

It should be pointed out that while the convolutional filter coefficients of the CNN are constrained to Pow2, the GAN was trained without this constraint, because the GAN architecture is not part of the final inference model. Only the GAN generator output is used during the training of CNNs.
In our case, the CNN is trained with 3456 normal and 3456 anomalous optical flow images, which are generated using the generator network of GAN.
We use binary cross entropy loss (also known as log loss) as the cost function. We stop the training when the Discriminator Loss ($L_D$) and Generator Loss ($L_G$) both fall below 0.01.



\label{ssec:pruning}



\begin{figure}[htb]
\vspace{-0.2in}
  \centering
    \includegraphics[scale=0.25]{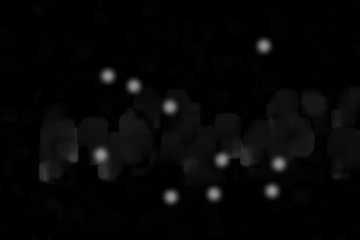}
    \caption{An example optical flow of a frame overlaid with 10 Gaussian blobs to simulate background motıon.}
    \label{res:modelnoise}
    \vspace{-0.1in}
\end{figure}
\begin{figure}[htb]
\vspace{-0.1in}
    \includegraphics[scale=0.5]{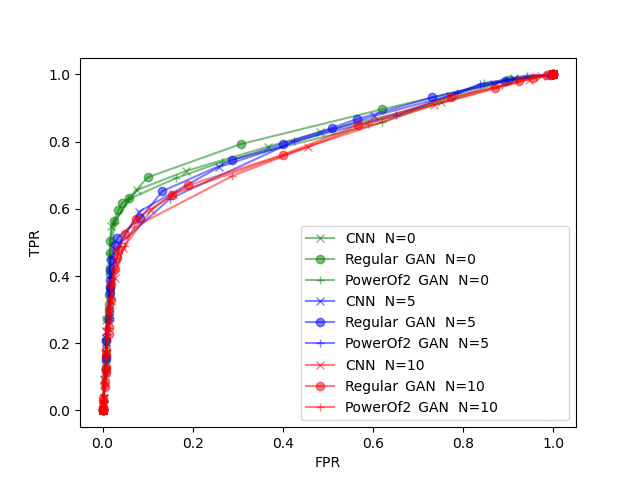}\vspace{-10pt}
    \caption{Comparison of a GAN trained CNN, a Pow2 GAN and a CNN at different noise levels. GAN trained CNN (dots) outperforms the CNN (other lines). GAN trained Pow2 GAN closely follows both the CNN and the GAN in terms of performance, while greatly reducing multiplications.}
    \label{res:noiseres}
    \vspace{-0.25in}
\end{figure}

{\em Robustness to background motion}:
We first compare our GAN improved CNN with a baseline CNN. We will perform this comparison at various noise levels.
In
Figure \ref{res:noiseres} we have the ROC curves corresponding to 9 different cases.
Background motion is modelled by adding random Gaussian blobs  onto the actual motion vector (mv) images. In Peds1 and Peds2 datasets there is no background motion due to leaves etc. That is why we added artificial motion vectors onto the optical flow images.
The number of blobs vary between 5 and 20. Their variance is 16 squared pixels. An example of a frame overlaid with Gaussian blobs is shown in Figure \ref{res:modelnoise}.
Due to the lack of space we plotted all the ROC curves on top of each other but the ROC curves of GAN improved CNNs (curves with big dots) have more area under them. Curves with green (blue) [red] dots have more area compared to other green, (blue) and [red] curves.
In Tables \ref{tab:ped1aucsavinghard} - \ref{tab:ped2aucsavingsoft} we present the Area Under Curve (AUC) values and percentage savings for different sets of both soft and hard denoising thresholds. All of the networks are also trained with artificial motion vector images generated by a GAN.
Denoising with hard thresholds 0.009 and 0.01 in the first and second layers, respectively, improves the AUC of GAN trained CNN while saving more than 9\% computational savings as shown in Table 1 in Peds1 dataset. In Pow2 network,
senoising with soft thresholds 0.009 and 0.01 leads to almost 20\% computational savings with better AUC as shown in the last row of Table 2. This network is also more robust to noise compared to all the other networks.  Soft thresholds of 0.003 and 0.02 provides even higher AUC values but computational savings are in the order of 7 \% for this network. As shown in Tables 1-4, when there is no noise computational savings can reach to almost 30\%  while detecting all of the abnormal events (trucks, skateboarders and bikers) in Peds1 and Peds2 datasets.
We also observe that layer denoising has a bigger advantage with Pow2 networks as compared to regular networks.

\begin{table*}[]
  \centering
\begin{tabular}{ |p{2.5cm}|p{2.21cm}|p{1cm}|p{1cm}|p{1cm}|p{1cm}|p{1cm}|p{1cm}|p{1cm}|p{1cm}|  }
 \hline
 & Threshold(s) & \multicolumn{2}{c|}{Noise = 0} & \multicolumn{2}{c|}{Noise = 5 blobs} &
         \multicolumn{2}{c|}{Noise = 10 blobs} & \multicolumn{2}{c|}{Noise = 20 blobs} \\
 \hline
 Network  & (Layer1, Layer2) & AUC & Saving & AUC & Saving & AUC & Saving & AUC & Saving \\
  \hline
  \hline
 Regular & N/A & 0.744 & 0.00\% & 0.727 & 0.00\% & 0.698 & 0.00\% & 0.641 & 0.00\% \\
 \hline
 \hline
 Denoising & (0.001, None) & 0.744 & 0.06\% & 0.727 & 0.13\% & 0.698 & 0.20\% & 0.641 & 0.33\% \\
 \hline
 Denoising & (0.001, 0.01) & 0.748 & 3.15\% & 0.727 & 3.24\% & 0.696 & 3.33\% & 0.635 & 3.50\% \\
 \hline
 Denoising & (0.002, None) & 0.744 & 2.05\% & 0.727 & 2.10\% & 0.697 & 2.16\% & 0.641 & 2.27\% \\
 \hline
 Denoising & (0.002, 0.01) & 0.748 & 5.13\% & 0.727 & 5.21\% & 0.695 & 5.28\% & 0.635 & 5.44\% \\
 \hline
 Denoising & (0.003, None) & 0.744 & 3.96\% & 0.727 & 3.97\% & 0.697 & 3.99\% & 0.641 & 4.03\% \\
 \hline
 Denoising & (0.003, 0.01) & 0.748 & 7.05\% & 0.727 & 7.08\% & 0.695 & 7.11\% & 0.635 & 7.20\% \\
 \hline
 Denoising & (0.005, None) & 0.744 & 4.17\% & 0.727 & 4.25\% & 0.697 & 4.33\% & 0.640 & 4.50\% \\
 \hline
 Denoising & 0.005, 0.01 & 0.748 & 7.26\% & 0.726 & 7.36\% & 0.695 & 7.46\% & 0.634 & 7.67\% \\
 \hline
 Denoising & (0.007, None) & 0.745 & 4.30\% & 0.727 & 4.44\% & 0.697 & 4.58\% & 0.639 & 4.86\% \\
 \hline
 Denoising & (0.007, 0.01) & 0.748 & 7.39\% & 0.727 & 7.55\% & 0.695 & 7.71\% & 0.633 & 8.03\% \\
 \hline
 Denoising & (0.009, None) & 0.744 & 6.19\% & 0.726 & 6.29\% & 0.696 & 6.39\% & 0.639 & 6.61\% \\
 \hline
 Denoising & (0.009, 0.01) & 0.748 & 9.28\% & 0.726 & 9.40\% & 0.694 & 9.53\% & 0.633 & 9.79\% \\
 \hline
 Denoising & (0.009, 0.02) & 0.749 & 10.04\% & 0.712 & 10.19\% & 0.673 & 10.34\% & 0.612 & 10.66\% \\
 \hline
 Denoising & (0.009, 0.03) & 0.754 & 11.31\% & 0.716 & 11.45\% & 0.656 & 11.60\% & 0.576 & 11.91\% \\
 \hline
 Denoising & (0.01, None) & 0.744 & 6.28\% & 0.727 & 6.40\% & 0.696 & 6.53\% & 0.639 & 6.80\% \\
 \hline
 Denoising & (0.02, None) & 0.744 & 10.26\% & 0.724 & 10.39\% & 0.691 & 10.53\% & 0.633 & 10.82\% \\
 \hline
 Denoising & (0.1, None) & 0.744 & 26.06\% & 0.710 & 26.16\% & 0.649 & 26.26\% & 0.543 & 26.46\% \\
 \hline
 Denoising & (0.1, 0.01) & 0.745 & 29.63\% & 0.707 & 29.77\% & 0.641 & 29.92\% & 0.528 & 30.22\% \\
 \hline
 \hline
 Power-of-2 (Pow2) without Denoising & N/A & 0.756 & 0.00\% & 0.741 & 0.00\% & 0.720 & 0.00\% & 0.687 & 0.00\% \\
 \hline
 \hline
 Denoising, Pow2 & (0.001, None) & 0.756 & 0.06\% & 0.741 & 0.08\% & 0.720 & 0.10\% & 0.687 & 0.14\% \\
 \hline
 Denoising, Pow2 & (0.001, 0.01) & 0.769 & 4.29\% & 0.752 & 4.37\% & 0.725 & 4.46\% & 0.666 & 4.62\% \\
 \hline
 Denoising, Pow2 & (0.002, None) & 0.756 & 0.16\% & 0.741 & 0.20\% & 0.720 & 0.24\% & 0.687 & 0.32\% \\
 \hline
 Denoising, Pow2 & (0.002, 0.01) & 0.769 & 4.39\% & 0.751 & 4.49\% & 0.725 & 4.59\% & 0.666 & 4.81\% \\
 \hline
 Denoising, Pow2 & (0.003, None) & 0.756 & 0.20\% & 0.741 & 0.26\% & 0.719 & 0.32\% & 0.686 & 0.46\% \\
 \hline
 Denoising, Pow2 & (0.003, 0.01) & 0.769 & 4.43\% & 0.751 & 4.55\% & 0.725 & 4.68\% & 0.665 & 4.94\% \\
 \hline
 Denoising, Pow2 & (0.005, None) & 0.756 & 0.32\% & 0.740 & 0.44\% & 0.718 & 0.56\% & 0.685 & 0.83\% \\
 \hline
 Denoising, Pow2 & (0.005, 0.01) & 0.769 & 4.54\% & 0.751 & 4.73\% & 0.724 & 4.92\% & 0.664 & 5.31\% \\
 \hline
 Denoising, Pow2 & (0.007, None) & 0.755 & 0.46\% & 0.739 & 0.67\% & 0.717 & 0.89\% & 0.684 & 1.33\% \\
 \hline
 Denoising, Pow2 & (0.007, 0.01) & 0.768 & 4.68\% & 0.750 & 4.96\% & 0.723 & 5.24\% & 0.662 & 5.82\% \\
 \hline
 Denoising, Pow2 & (0.009, None) & 0.758 & 0.91\% & 0.745 & 1.27\% & 0.726 & 1.62\% & 0.691 & 2.32\% \\
 \hline
 Denoising, Pow2 & (0.009, 0.01) & 0.768 & 5.14\% & 0.752 & 5.56\% & 0.727 & 5.98\% & 0.663 & 6.81\% \\
 \hline
 Denoising, Pow2 & (0.009, 0.02) & 0.768 & 7.92\% & 0.756 & 8.29\% & 0.740 & 8.65\% & 0.703 & 9.37\% \\
 \hline
 Denoising, Pow2 & (0.009, 0.03) & 0.746 & 8.40\% & 0.710 & 8.76\% & 0.683 & 9.12\% & 0.632 & 9.83\% \\
 \hline
 Denoising, Pow2 & (0.01, None) & 0.755 & 2.84\% & 0.744 & 3.19\% & 0.724 & 3.54\% & 0.688 & 4.22\% \\
 \hline
 Denoising, Pow2 & (0.02, None) & 0.759 & 14.43\% & 0.744 & 14.66\% & 0.722 & 14.88\% & 0.686 & 15.32\% \\
 \hline
 Denoising, Pow2 & (0.1, None) & 0.756 & 28.00\% & 0.722 & 28.04\% & 0.688 & 28.07\% & 0.644 & 28.15\% \\
 \hline
 Denoising, Pow2 & (0.1, 0.01) & 0.766 & 33.18\% & 0.742 & 33.29\% & 0.699 & 33.40\% & 0.631 & 33.60\% \\
 \hline
\end{tabular}
\caption{[UCSD Pedestrians 1 Dataset] Observed Area Under Curve (AUC) of ROCs and percentage savings on different optimizations using  hard thresholding. Denoising not only improves AUC in both regular and Pow2 networks but also provides 9.28 and 28.0 \% savings in regular and Pow2 networks, respectively. Soft thresholding provides robustness against noise as shown in Table 2.}
  \label{tab:ped1aucsavinghard}
  \vspace{-0.1in}
\end{table*}
\vspace{-0.1in}
\begin{table*}[]
  \centering
  \vspace{-0.1in}
\begin{tabular}{ |p{2.6cm}|p{2.2cm}|p{1cm}|p{1cm}|p{1cm}|p{1cm}|p{1cm}|p{1cm}|p{1cm}|p{1cm}|  }
 \hline
 & Threshold(s) & \multicolumn{2}{c|}{Noise = 0} & \multicolumn{2}{c|}{Noise = 5 blobs} &
         \multicolumn{2}{c|}{Noise = 10 blobs} & \multicolumn{2}{c|}{Noise = 20 blobs} \\
 \hline
 Network & (Layer1, Layer2) & AUC & Saving & AUC & Saving & AUC & Saving & AUC & Saving \\
  \hline
  \hline
 Regular & N/A & 0.744 & 0.00\% & 0.727 & 0.00\% & 0.698 & 0.00\% & 0.687 & 0.00\% \\
 \hline
 \hline
 Denoising & (0.001, None) & 0.745 & 2.05\% & 0.727 & 2.10\% & 0.697 & 2.16\% & 0.686 & 0.32\% \\
 \hline
 Denoising & (0.001, 0.01) & 0.753 & 5.84\% & 0.717 & 5.95\% & 0.670 & 6.06\% & 0.696 & 7.37\% \\
 \hline
 Denoising & (0.002, None) & 0.745 & 4.06\% & 0.727 & 4.11\% & 0.696 & 4.16\% & 0.686 & 0.64\% \\
 \hline
 Denoising & (0.002, 0.01) & 0.753 & 7.89\% & 0.717 & 7.99\% & 0.668 & 8.10\% & 0.696 & 7.70\% \\
 \hline
 Denoising & (0.003, None) & 0.746 & 4.22\% & 0.727 & 4.33\% & 0.696 & 4.44\% & 0.685 & 1.06\% \\
 \hline
 Denoising & (0.003, 0.01) & 0.753 & 8.09\% & 0.716 & 8.25\% & 0.667 & 8.42\% & 0.695 & 8.12\% \\
 \hline
 Denoising & (0.005, None) & 0.746 & 6.28\% & 0.727 & 6.40\% & 0.694 & 6.53\% & 0.686 & 4.22\% \\
 \hline
 Denoising & 0.005, 0.01 & 0.754 & 10.46\% & 0.721 & 10.62\% & 0.670 & 10.78\% & 0.691 & 11.30\% \\
 \hline
 Denoising & (0.007, None) & 0.749 & 6.83\% & 0.726 & 7.04\% & 0.690 & 7.26\% & 0.683 & 11.28\% \\
 \hline
 Denoising & (0.007, 0.01) & 0.755 & 11.03\% & 0.715 & 11.29\% & 0.659 & 11.55\% & 0.690 & 18.37\% \\
 \hline
 Denoising & (0.009, None) & 0.748 & 8.73\% & 0.726 & 8.93\% & 0.690 & 9.12\% & 0.681 & 13.95\% \\
 \hline
 Denoising & (0.009, 0.01) & 0.756 & 12.98\% & 0.713 & 13.22\% & 0.654 & 13.46\% & 0.687 & 21.05\% \\
 \hline
 Denoising & (0.009, 0.02) & 0.759 & 15.36\% & 0.717 & 15.60\% & 0.657 & 15.83\% & 0.454 & 21.92\% \\
 \hline
 Denoising & (0.009, 0.03) & 0.500 & 16.71\% & 0.500 & 16.91\% & 0.500 & 17.10\% & 0.500 & 21.93\% \\
 \hline
 Denoising & (0.01, None) & 0.748 & 10.26\% & 0.726 & 10.39\% & 0.689 & 10.53\% & 0.683 & 15.32\% \\
 \hline
 Denoising & (0.02, None) & 0.756 & 16.58\% & 0.727 & 16.82\% & 0.681 & 17.06\% & 0.671 & 25.93\% \\
 \hline
 Denoising & (0.1, None) & 0.500 & 31.91\% & 0.500 & 31.91\% & 0.500 & 31.91\% & 0.577 & 30.55\% \\
 \hline
 Denoising & (0.1, 0.01) & 0.500 & 39.89\% & 0.500 & 39.89\% & 0.500 & 39.89\% & 0.467 & 38.30\% \\
 \hline
 \hline
 Power-of-2 (Pow2) without Denoising & N/A & 0.756 & 0.00\% & 0.741 & 0.00\% & 0.720 & 0.00\% & 0.687 & 0.00\% \\
 \hline
 \hline
 Denoising, Pow2 & (0.001, None) & 0.756 & 0.16\% & 0.741 & 0.20\% & 0.720 & 0.24\% & 0.686 & 0.32\% \\
 \hline
 Denoising, Pow2 & (0.001, 0.01) & 0.768 & 7.17\% & 0.753 & 7.21\% & 0.733 & 7.26\% & 0.696 & 7.37\% \\
 \hline
 Denoising, Pow2 & (0.002, None) & 0.756 & 0.27\% & 0.741 & 0.36\% & 0.719 & 0.45\% & 0.686 & 0.64\% \\
 \hline
 Denoising, Pow2 & (0.002, 0.01) & 0.769 & 7.28\% & 0.754 & 7.38\% & 0.734 & 7.48\% & 0.696 & 7.70\% \\
 \hline
 Denoising, Pow2 & (0.003, None) & 0.756 & 0.39\% & 0.741 & 0.55\% & 0.719 & 0.72\% & 0.685 & 1.06\% \\
 \hline
 Denoising, Pow2 & (0.003, 0.01) & 0.769 & 7.40\% & 0.755 & 7.58\% & 0.733 & 7.75\% & 0.695 & 8.12\% \\
 \hline
 Denoising, Pow2 & (0.005, None) & 0.756 & 2.84\% & 0.743 & 3.19\% & 0.722 & 3.54\% & 0.686 & 4.22\% \\
 \hline
 Denoising, Pow2 & (0.005, 0.01) & 0.761 & 9.85\% & 0.751 & 10.22\% & 0.731 & 10.58\% & 0.691 & 11.30\% \\
 \hline
 Denoising, Pow2 & (0.007, None) & 0.757 & 10.40\% & 0.742 & 10.62\% & 0.720 & 10.84\% & 0.683 & 11.28\% \\
 \hline
 Denoising, Pow2 & (0.007, 0.01) & 0.765 & 17.42\% & 0.752 & 17.66\% & 0.730 & 17.90\% & 0.690 & 18.37\% \\
 \hline
 Denoising, Pow2 & (0.009, None) & 0.757 & 12.96\% & 0.741 & 13.21\% & 0.719 & 13.47\% & 0.681 & 13.95\% \\
 \hline
 Denoising, Pow2 & (0.009, 0.01) & 0.764 & 19.99\% & 0.750 & 20.26\% & 0.727 & 20.53\% & 0.687 & 21.05\% \\
 \hline
 Denoising, Pow2 & (0.009, 0.02) & 0.459 & 20.92\% & 0.459 & 21.18\% & 0.459 & 21.43\% & 0.454 & 21.92\% \\
 \hline
 Denoising, Pow2 & (0.009, 0.03) & 0.500 & 20.94\% & 0.500 & 21.19\% & 0.500 & 21.44\% & 0.500 & 21.93\% \\
 \hline
 Denoising, Pow2 & (0.01, None) & 0.758 & 14.43\% & 0.744 & 14.66\% & 0.721 & 14.88\% & 0.683 & 15.32\% \\
 \hline
 Denoising, Pow2 & (0.02, None) & 0.758 & 25.85\% & 0.741 & 25.87\% & 0.716 & 25.89\% & 0.671 & 25.93\% \\
 \hline
 Denoising, Pow2 & (0.1, None) & 0.695 & 30.27\% & 0.666 & 30.35\% & 0.628 & 30.42\% & 0.577 & 30.55\% \\
 \hline
 Denoising, Pow2 & (0.1, 0.01) & 0.626 & 37.93\% & 0.567 & 38.03\% & 0.523 & 38.13\% & 0.467 & 38.30\% \\
 \hline
\end{tabular}
  \caption{[UCSD Pedestrians 1 Dataset] Observed Area Under Curve (AUC) of ROCs and percentage savings on different optimizations using  soft thresholding, which provides better AUC results in both regular and Pow2 networks under no-noise and noisy conditions.}
  \label{tab:ped1aucsavingsoft}
\end{table*}
\vspace{-0.1in}
\begin{table*}[]
  \centering
\begin{tabular}{ |p{2.6cm}|p{2.2cm}|p{1cm}|p{1cm}|p{1cm}|p{1cm}|p{1cm}|p{1cm}|p{1cm}|p{1cm}|  }
 \hline
 & Threshold(s) & \multicolumn{2}{c|}{Noise = 0} & \multicolumn{2}{c|}{Noise = 5 blobs} &
         \multicolumn{2}{c|}{Noise = 10 blobs} & \multicolumn{2}{c|}{Noise = 20 blobs} \\
 \hline
 Network & (Layer1, Layer2) & AUC & Saving & AUC & Saving & AUC & Saving & AUC & Saving \\
  \hline
  \hline
 Regular & N/A & 0.837 & 0.00\% & 0.811 & 0.00\% & 0.793 & 0.00\% & 0.751 & 0.00\% \\
 \hline
 \hline
 Denoising & (0.001, None) & 0.837 & 0.06\% & 0.811 & 0.09\% & 0.793 & 0.13\% & 0.751 & 0.19\% \\
 \hline
 Denoising & (0.001, 0.01) & 0.838 & 3.14\% & 0.812 & 3.19\% & 0.794 & 3.23\% & 0.751 & 3.31\% \\
 \hline
 Denoising & (0.002, None) & 0.836 & 2.07\% & 0.811 & 2.09\% & 0.793 & 2.11\% & 0.750 & 2.15\% \\
 \hline
 Denoising & (0.002, 0.01) & 0.838 & 5.16\% & 0.811 & 5.18\% & 0.793 & 5.21\% & 0.750 & 5.27\% \\
 \hline
 Denoising & (0.003, None) & 0.837 & 3.94\% & 0.811 & 3.94\% & 0.793 & 3.93\% & 0.750 & 3.93\% \\
 \hline
 Denoising & (0.003, 0.01) & 0.838 & 7.02\% & 0.811 & 7.03\% & 0.793 & 7.03\% & 0.750 & 7.04\% \\
 \hline
 Denoising & (0.005, None) & 0.836 & 4.15\% & 0.811 & 4.19\% & 0.793 & 4.22\% & 0.750 & 4.29\% \\
 \hline
 Denoising & 0.005, 0.01 & 0.837 & 7.24\% & 0.811 & 7.28\% & 0.793 & 7.32\% & 0.749 & 7.41\% \\
 \hline
 Denoising & (0.007, None) & 0.836 & 4.31\% & 0.811 & 4.38\% & 0.793 & 4.44\% & 0.750 & 4.57\% \\
 \hline
 Denoising & (0.007, 0.01) & 0.837 & 7.40\% & 0.811 & 7.47\% & 0.793 & 7.55\% & 0.750 & 7.69\% \\
 \hline
 Denoising & (0.009, None) & 0.836 & 6.15\% & 0.810 & 6.20\% & 0.792 & 6.24\% & 0.749 & 6.32\% \\
 \hline
 Denoising & (0.009, 0.01) & 0.837 & 9.24\% & 0.811 & 9.29\% & 0.792 & 9.34\% & 0.749 & 9.45\% \\
 \hline
 Denoising & (0.009, 0.02) & 0.819 & 9.98\% & 0.792 & 10.05\% & 0.774 & 10.11\% & 0.741 & 10.24\% \\
 \hline
 Denoising & (0.009, 0.03) & 0.821 & 11.27\% & 0.797 & 11.32\% & 0.779 & 11.38\% & 0.744 & 11.50\% \\
 \hline
 Denoising & (0.01, None) & 0.836 & 6.25\% & 0.810 & 6.30\% & 0.792 & 6.36\% & 0.749 & 6.47\% \\
 \hline
 Denoising & (0.02, None) & 0.836 & 10.20\% & 0.808 & 10.26\% & 0.788 & 10.32\% & 0.745 & 10.45\% \\
 \hline
 Denoising & (0.1, None) & 0.809 & 26.05\% & 0.786 & 26.09\% & 0.763 & 26.14\% & 0.729 & 26.23\% \\
 \hline
 Denoising & (0.1, 0.01) & 0.807 & 29.60\% & 0.787 & 29.66\% & 0.762 & 29.72\% & 0.728 & 29.84\% \\
 \hline
 \hline
 Power-of-2 (Pow2) without Denoising & N/A & 0.811 & 0.00\% & 0.800 & 0.00\% & 0.783 & 0.00\% & 0.751 & 0.00\% \\
 \hline
 \hline
 Denoising, Pow2 & (0.001, None) & 0.811 & 0.06\% & 0.800 & 0.07\% & 0.783 & 0.07\% & 0.751 & 0.09\% \\
 \hline
 Denoising, Pow2 & (0.001, 0.01) & 0.824 & 4.30\% & 0.805 & 4.33\% & 0.789 & 4.37\% & 0.756 & 4.44\% \\
 \hline
 Denoising, Pow2 & (0.002, None) & 0.811 & 0.14\% & 0.800 & 0.16\% & 0.783 & 0.17\% & 0.751 & 0.20\% \\
 \hline
 Denoising, Pow2 & (0.002, 0.01) & 0.824 & 4.38\% & 0.805 & 4.43\% & 0.788 & 4.47\% & 0.756 & 4.56\% \\
 \hline
 Denoising, Pow2 & (0.003, None) & 0.811 & 0.19\% & 0.799 & 0.21\% & 0.783 & 0.23\% & 0.751 & 0.28\% \\
 \hline
 Denoising, Pow2 & (0.003, 0.01) & 0.824 & 4.43\% & 0.804 & 4.48\% & 0.788 & 4.53\% & 0.756 & 4.63\% \\
 \hline
 Denoising, Pow2 & (0.005, None) & 0.811 & 0.28\% & 0.799 & 0.33\% & 0.782 & 0.38\% & 0.750 & 0.49\% \\
 \hline
 Denoising, Pow2 & (0.005, 0.01) & 0.824 & 4.52\% & 0.805 & 4.59\% & 0.788 & 4.68\% & 0.754 & 4.84\% \\
 \hline
 Denoising, Pow2 & (0.007, None) & 0.811 & 0.42\% & 0.798 & 0.51\% & 0.781 & 0.60\% & 0.747 & 0.79\% \\
 \hline
 Denoising, Pow2 & (0.007, 0.01) & 0.823 & 4.65\% & 0.803 & 4.77\% & 0.786 & 4.89\% & 0.752 & 5.14\% \\
 \hline
 Denoising, Pow2 & (0.009, None) & 0.817 & 0.98\% & 0.808 & 1.14\% & 0.793 & 1.30\% & 0.764 & 1.61\% \\
 \hline
 Denoising, Pow2 & (0.009, 0.01) & 0.823 & 5.21\% & 0.808 & 5.40\% & 0.795 & 5.59\% & 0.767 & 5.97\% \\
 \hline
 Denoising, Pow2 & (0.009, 0.02) & 0.815 & 7.99\% & 0.802 & 8.15\% & 0.796 & 8.31\% & 0.777 & 8.63\% \\
 \hline
 Denoising, Pow2 & (0.009, 0.03) & 0.813 & 8.46\% & 0.788 & 8.62\% & 0.781 & 8.78\% & 0.740 & 9.09\% \\
 \hline
 Denoising, Pow2 & (0.01, None) & 0.801 & 2.92\% & 0.791 & 3.07\% & 0.777 & 3.22\% & 0.756 & 3.52\% \\
 \hline
 Denoising, Pow2 & (0.02, None) & 0.815 & 14.45\% & 0.804 & 14.56\% & 0.787 & 14.66\% & 0.756 & 14.87\% \\
 \hline
 Denoising, Pow2 & (0.1, None) & 0.805 & 27.99\% & 0.773 & 28.00\% & 0.746 & 28.01\% & 0.711 & 28.04\% \\
 \hline
 Denoising, Pow2 & (0.1, 0.01) & 0.769 & 33.20\% & 0.754 & 33.25\% & 0.735 & 33.30\% & 0.717 & 33.40\% \\
 \hline
\end{tabular}
  \caption{[UCSD Pedestrians 2 Dataset] Observed Area Under Curve (AUC) of ROCs and percentage savings on different optimizations using  hard thresholding. This dataset is smaller than Pedestrian 1 dataset.}
  \label{tab:ped2aucsavinghard}
\end{table*}
\begin{table*}[]
  \centering
\begin{tabular}{ |p{2.6cm}|p{2.2cm}|p{1cm}|p{1cm}|p{1cm}|p{1cm}|p{1cm}|p{1cm}|p{1cm}|p{1cm}|  }
 \hline
 & Threshold(s) & \multicolumn{2}{c|}{Noise = 0} & \multicolumn{2}{c|}{Noise = 5 blobs} &
         \multicolumn{2}{c|}{Noise = 10 blobs} & \multicolumn{2}{c|}{Noise = 20 blobs} \\
 \hline
 Network  & (Layer1, Layer2) & AUC & Saving & AUC & Saving & AUC & Saving & AUC & Saving \\
  \hline
  \hline
 Regular & N/A & 0.837 & 0.00\% & 0.811 & 0.00\% & 0.793 & 0.00\% & 0.751 & 0.00\% \\
 \hline
 \hline
 Denoising & (0.001, None) & 0.835 & 2.07\% & 0.810 & 2.09\% & 0.793 & 2.11\% & 0.750 & 2.15\% \\
 \hline
 Denoising & (0.001, 0.01) & 0.817 & 5.86\% & 0.792 & 5.91\% & 0.777 & 5.95\% & 0.747 & 6.03\% \\
 \hline
 Denoising & (0.002, None) & 0.834 & 4.03\% & 0.810 & 4.05\% & 0.792 & 4.07\% & 0.750 & 4.11\% \\
 \hline
 Denoising & (0.002, 0.01) & 0.818 & 7.85\% & 0.792 & 7.89\% & 0.777 & 7.93\% & 0.747 & 8.02\% \\
 \hline
 Denoising & (0.003, None) & 0.833 & 4.21\% & 0.809 & 4.26\% & 0.791 & 4.31\% & 0.750 & 4.41\% \\
 \hline
 Denoising & (0.003, 0.01) & 0.815 & 8.06\% & 0.790 & 8.13\% & 0.775 & 8.21\% & 0.745 & 8.36\% \\
 \hline
 Denoising & (0.005, None) & 0.831 & 6.25\% & 0.807 & 6.30\% & 0.790 & 6.36\% & 0.750 & 6.47\% \\
 \hline
 Denoising & 0.005, 0.01 & 0.826 & 10.41\% & 0.801 & 10.48\% & 0.785 & 10.55\% & 0.753 & 10.68\% \\
 \hline
 Denoising & (0.007, None) & 0.821 & 6.77\% & 0.800 & 6.88\% & 0.784 & 6.98\% & 0.748 & 7.19\% \\
 \hline
 Denoising & (0.007, 0.01) & 0.801 & 10.98\% & 0.779 & 11.10\% & 0.765 & 11.22\% & 0.739 & 11.46\% \\
 \hline
 Denoising & (0.009, None) & 0.827 & 8.67\% & 0.803 & 8.77\% & 0.785 & 8.87\% & 0.747 & 9.07\% \\
 \hline
 Denoising & (0.009, 0.01) & 0.795 & 12.93\% & 0.772 & 13.04\% & 0.758 & 13.16\% & 0.732 & 13.39\% \\
 \hline
 Denoising & (0.009, 0.02) & 0.825 & 15.30\% & 0.795 & 15.42\% & 0.772 & 15.53\% & 0.737 & 15.76\% \\
 \hline
 Denoising & (0.009, 0.03) & 0.500 & 16.65\% & 0.500 & 16.75\% & 0.500 & 16.85\% & 0.500 & 17.04\% \\
 \hline
 Denoising & (0.01, None) & 0.826 & 10.20\% & 0.803 & 10.26\% & 0.785 & 10.32\% & 0.748 & 10.45\% \\
 \hline
 Denoising & (0.02, None) & 0.809 & 16.65\% & 0.791 & 16.75\% & 0.776 & 16.85\% & 0.749 & 17.04\% \\
 \hline
 Denoising & (0.1, None) & 0.500 & 31.91\% & 0.500 & 31.91\% & 0.500 & 31.91\% & 0.500 & 31.91\% \\
 \hline
 Denoising & (0.1, 0.01) & 0.500 & 39.89\% & 0.500 & 39.89\% & 0.500 & 39.89\% & 0.500 & 39.89\% \\
 \hline
 \hline
 Power-of-2 (Pow2) without Denoising & N/A & 0.811 & 0.00\% & 0.800 & 0.00\% & 0.783 & 0.00\% & 0.751 & 0.00\% \\
 \hline
 \hline
 Denoising, Pow2 & (0.001, None) & 0.810 & 0.14\% & 0.799 & 0.16\% & 0.783 & 0.17\% & 0.751 & 0.20\% \\
 \hline
 Denoising, Pow2 & (0.001, 0.01) & 0.807 & 7.15\% & 0.794 & 7.17\% & 0.789 & 7.19\% & 0.775 & 7.23\% \\
 \hline
 Denoising, Pow2 & (0.002, None) & 0.809 & 0.25\% & 0.798 & 0.28\% & 0.782 & 0.32\% & 0.751 & 0.39\% \\
 \hline
 Denoising, Pow2 & (0.002, 0.01) & 0.807 & 7.26\% & 0.795 & 7.30\% & 0.789 & 7.34\% & 0.777 & 7.42\% \\
 \hline
 Denoising, Pow2 & (0.003, None) & 0.808 & 0.35\% & 0.798 & 0.42\% & 0.781 & 0.49\% & 0.750 & 0.63\% \\
 \hline
 Denoising, Pow2 & (0.003, 0.01) & 0.808 & 7.36\% & 0.795 & 7.44\% & 0.789 & 7.51\% & 0.780 & 7.66\% \\
 \hline
 Denoising, Pow2 & (0.005, None) & 0.801 & 2.92\% & 0.792 & 3.07\% & 0.777 & 3.22\% & 0.753 & 3.52\% \\
 \hline
 Denoising, Pow2 & (0.005, 0.01) & 0.809 & 9.95\% & 0.794 & 10.10\% & 0.785 & 10.26\% & 0.780 & 10.57\% \\
 \hline
 Denoising, Pow2 & (0.007, None) & 0.808 & 10.38\% & 0.799 & 10.49\% & 0.782 & 10.59\% & 0.752 & 10.79\% \\
 \hline
 Denoising, Pow2 & (0.007, 0.01) & 0.804 & 17.42\% & 0.789 & 17.53\% & 0.778 & 17.64\% & 0.776 & 17.86\% \\
 \hline
 Denoising, Pow2 & (0.009, None) & 0.802 & 12.88\% & 0.794 & 13.00\% & 0.777 & 13.13\% & 0.749 & 13.37\% \\
 \hline
 Denoising, Pow2 & (0.009, 0.01) & 0.793 & 19.91\% & 0.778 & 20.05\% & 0.770 & 20.18\% & 0.773 & 20.44\% \\
 \hline
 Denoising, Pow2 & (0.009, 0.02) & 0.516 & 20.84\% & 0.483 & 20.97\% & 0.512 & 21.09\% & 0.507 & 21.34\% \\
 \hline
 Denoising, Pow2 & (0.009, 0.03) & 0.500 & 20.85\% & 0.500 & 20.98\% & 0.500 & 21.11\% & 0.500 & 21.35\% \\
 \hline
 Denoising, Pow2 & (0.01, None) & 0.810 & 14.45\% & 0.799 & 14.56\% & 0.782 & 14.66\% & 0.754 & 14.87\% \\
 \hline
 Denoising, Pow2 & (0.02, None) & 0.804 & 25.84\% & 0.793 & 25.84\% & 0.776 & 25.84\% & 0.748 & 25.84\% \\
 \hline
 Denoising, Pow2 & (0.1, None) & 0.618 & 30.35\% & 0.612 & 30.38\% & 0.615 & 30.41\% & 0.573 & 30.47\% \\
 \hline
 Denoising, Pow2 & (0.1, 0.01) & 0.569 & 38.01\% & 0.568 & 38.06\% & 0.553 & 38.10\% & 0.515 & 38.20\% \\
 \hline
\end{tabular}
  \caption{[UCSD Pedestrians 2 Dataset] Observed Area Under Curve (AUC) of ROCs and percentage savings on different optimizations using soft thresholding, which provides robustness against noise as shown in the last row. AUC values of the soft thresholded network is  higher than corresponding cases when the is noise is high while providing about 10\% savings. }
  \label{tab:ped2aucsavingsoft}
\end{table*}

\section{Conclusion}
\label{sec:conclusion}
\vspace{-0.1in}
In this paper, we developed a computationally efficient anomaly detection algorithm using motion vector images and Pow2 arithmetic.
Our deep neural network inference algorithm is about 10\% to 25\% faster than the corresponding powers-of-two networks while detecting all the anomalous events with almost the same AUC. We reduced the complexity of the network by denoising the first two output layers. As a result, the anomaly detection scheme can be implemented  in real-time on low power devices.

The resulting system turns out to be more robust to background motion because the denoising forces small valued intermediate outputs to zero. This process eliminates the contribution of small motion vectors to the final result in Pow2 networks. Furthermore, we augmented the training data set of the Pow2 network with a GAN generated images to improve the anomaly detection rate.

Future work will include enhancing the denoised powers-of-two networks with other complexity reduction techniques such as network pruning or conditional execution \cite{DBLP:journals/corr/abs-1710-09282,DBLP:conf/date/PandaSR16,biasielli1}.



\bibliographystyle{IEEEbib}
\bibliography{strings,refs}

\begin{thebibliography}{10}

\bibitem{DBLP:conf/nips/GoodfellowPMXWOCB14}
I.~J. Goodfellow, J.~Pouget{-}Abadie, M.~Mirza, B.~Xu, D.~Warde{-}Farley,
  S.~Ozair, A.~C. Courville, and Y.~Bengio,
\newblock ``Generative adversarial nets,''
\newblock in {\em Proceedings of NIPS 2014}, 2014, pp. 2672--2680.

\bibitem{DBLP:journals/corr/abs-1802-06222}
H.~Zenati, C.~Sheng Foo, B.~Lecouat, G.~Manek, and V.~Ramaseshan Chandrasekhar,
\newblock ``Efficient gan-based anomaly detection,''
\newblock {\em CoRR}, vol. abs/1802.06222, 2018.

\bibitem{DBLP:conf/icip/RavanbakhshNSMR17}
M.~Ravanbakhsh, M.~Nabi, E.~Sangineto, L.~Marcenaro, C.~S. Regazzoni, and
  N.~Sebe,
\newblock ``Abnormal event detection in videos using generative adversarial
  nets,''
\newblock in {\em 2017 {IEEE} {ICIP} 2017}, 2017, pp. 1577--1581.

\bibitem{DBLP:conf/dac/TannHBR17}
H.~Tann, S.~Hashemi, R.~I. Bahar, and S.~Reda,
\newblock ``Hardware-software codesign of accurate, multiplier-free deep neural
  networks,''
\newblock in {\em Proceedings of {DAC} 2017}, 2017, pp. 28:1--28:6.

\bibitem{DBLP:conf/dac/DingLCMB19}
R.~Ding, Z.~Liu, Ting{-}Wu Chin, D.~Marculescu, and R.~D.~(Shawn) Blanton,
\newblock ``Flightnns: Lightweight quantized deep neural networks for fast and
  accurate inference,''
\newblock in {\em Proceedings of {DAC} 2019}, 2019, p. 200.

\bibitem{DBLP:journals/tit/KrimTMD99}
H.~Krim, D.~Tucker, S.~Mallat, and D.~L. Donoho,
\newblock ``On denoising and best signal representation,''
\newblock {\em {IEEE} Trans. Information Theory}, vol. 45, no. 7, pp.
  2225--2238, 1999.

\bibitem{DBLP:conf/wacv/MousaviMPCM15}
H.~Mousavi, S.~Mohammadi, A.~Perina, R.~Chellali, and V.~Murino,
\newblock ``Analyzing tracklets for the detection of abnormal crowd behavior,''
\newblock in {\em 2015 {IEEE} {WACV} 2015}, 2015, pp. 148--155.

\bibitem{DBLP:conf/cvpr/MehranOS09}
R.~Mehran, A.~Oyama, and M.~Shah,
\newblock ``Abnormal crowd behavior detection using social force model,''
\newblock in {\em 2009 {IEEE} {CVPR} 2009}, 2009, pp. 935--942.

\bibitem{DBLP:conf/cvpr/MahadevanLBV10}
V.~Mahadevan, W.~Li, V.~Bhalodia, and N.~Vasconcelos,
\newblock ``Anomaly detection in crowded scenes,''
\newblock in {\em {IEEE} {CVPR} 2010}, 2010, pp. 1975--1981.

\bibitem{DBLP:conf/cvpr/CongYL11}
Y.~Cong, J.~Yuan, and J.~Liu,
\newblock ``Sparse reconstruction cost for abnormal event detection,''
\newblock in {\em {IEEE} {CVPR} 2011}, 2011, pp. 3449--3456.

\bibitem{DBLP:conf/cvpr/KimG09}
J.~Kim and K.~Grauman,
\newblock ``Observe locally, infer globally: {A} space-time {MRF} for detecting
  abnormal activities with incremental updates,''
\newblock in {\em 2009 {IEEE} {CVPR} 2009}, 2009, pp. 2921--2928.

\bibitem{DBLP:books/sp/13/RaghavendraCBSM13}
R.~Raghavendra, M.~Cristani, A.~{Del Bue}, E.~Sangineto, and V.~Murino,
\newblock ``Anomaly detection in crowded scenes: {A} novel framework based on
  swarm optimization and social force modeling,''
\newblock in {\em Modeling, Simulation and Visual Analysis of Crowds - {A}
  Multidisciplinary Perspective}, pp. 383--411. 2013.

\bibitem{DBLP:conf/avss/RabieeHMKNM16}
H.~R. Rabiee, J.~Haddadnia, H.~Mousavi, M.~Kalantarzadeh, M.~Nabi, and
  V.~Murino,
\newblock ``Novel dataset for fine-grained abnormal behavior understanding in
  crowd,''
\newblock in {\em {IEEE} {AVSS} 2016}, 2016, pp. 95--101.

\bibitem{DBLP:conf/cvpr/SaligramaC12}
V.~Saligrama and Z.~Chen,
\newblock ``Video anomaly detection based on local statistical aggregates,''
\newblock in {\em {IEEE} {CVPR} 2012}, 2012, pp. 2112--2119.

\bibitem{DBLP:journals/mlc/RabieeMNR18}
H.~R. Rabiee, H.~Mousavi, M.~Nabi, and M.~Ravanbakhsh,
\newblock ``Detection and localization of crowd behavior using a novel
  tracklet-based model,''
\newblock {\em Int. J. Machine Learning {\&} Cybernetics}, vol. 9, no. 12, pp.
  1999--2010, 2018.

\bibitem{DBLP:journals/itiis/HuangWSFK16}
X.~Huang, W.~Wang, G.~Shen, X.~Feng, and X.~Kong,
\newblock ``Crowd activity classification using category constrained correlated
  topic model,''
\newblock {\em {TIIS}}, vol. 10, no. 11, pp. 5530--5546, 2016.

\bibitem{DBLP:conf/wacv/RavanbakhshNMSS18}
M.~Ravanbakhsh, M.~Nabi, H.~Mousavi, E.~Sangineto, and N.~Sebe,
\newblock ``Plug-and-play {CNN} for crowd motion analysis: An application in
  abnormal event detection,''
\newblock in {\em {IEEE} {WACV} 2018}, 2018, pp. 1689--1698.

\bibitem{DBLP:journals/cviu/SabokrouFFMK18}
M.~Sabokrou, M.~Fayyaz, M.~Fathy, Z.~Moayed, and R.~Klette,
\newblock ``Deep-anomaly: Fully convolutional neural network for fast anomaly
  detection in crowded scenes,''
\newblock {\em Computer Vision and Image Understanding}, vol. 172, pp. 88--97,
  2018.

\bibitem{DBLP:journals/cviu/XuYRS17}
D.~Xu, Y.~Yan, E.~Ricci, and N.~Sebe,
\newblock ``Detecting anomalous events in videos by learning deep
  representations of appearance and motion,''
\newblock {\em Computer Vision and Image Understanding}, vol. 156, pp.
  117--127, 2017.

\bibitem{DSET:UCSD/PEDS}
``Peds dataset,'' http://www.svcl.ucsd.edu/projects/anomaly.

\bibitem{DBLP:conf/scia/Farneback03}
G.~Farneb{\"{a}}ck,
\newblock ``Two-frame motion estimation based on polynomial expansion,''
\newblock in {\em {SCIA} 2003, Halmstad, Sweden}, 2003, pp. 363--370.

\bibitem{DBLP:conf/cvpr/PathakKDDE16}
D.~Pathak, P.~Kr{\"{a}}henb{\"{u}}hl, J.~Donahue, T.~Darrell, and A.~A. Efros,
\newblock ``Context encoders: Feature learning by inpainting,''
\newblock in {\em {IEEE} {CVPR} 2016}, 2016, pp. 2536--2544.

\bibitem{DBLP:conf/cvpr/IsolaZZE17}
P.~Isola, Jun{-}Yan Zhu, T.~Zhou, and A.~A. Efros,
\newblock ``Image-to-image translation with conditional adversarial networks,''
\newblock in {\em {IEEE} {CVPR} 2017}, 2017, pp. 5967--5976.

\bibitem{RUDIN1992259}
L.~I. Rudin, S.~Osher, and E.~Fatemi,
\newblock ``Nonlinear total variation based noise removal algorithms,''
\newblock {\em Physica D: Nonlinear Phenomena}, vol. 60, no. 1, pp. 259 -- 268,
  1992.

\bibitem{10.2307/2346178}
R.~Tibshirani,
\newblock ``Regression shrinkage and selection via the lasso,''
\newblock {\em Journal of the Royal Statistical Society. Series B
  (Methodological)}, vol. 58, no. 1, pp. 267--288, 1996.

\bibitem{DBLP:journals/spm/CetinT15}
A.~E. {\c{C}}etin and M.~Tofighi,
\newblock ``Projection-based wavelet denoising [lecture notes],''
\newblock {\em {IEEE} Signal Process. Mag.}, vol. 32, no. 5, pp. 120--124,
  2015.

\bibitem{DBLP:conf/icip/TofighiKC14}
M.~Tofighi, K.~K{\"{o}}se, and A.~Enis {\c{C}}etin,
\newblock ``Denoising using projections onto the epigraph set of convex cost
  functions,''
\newblock in {\em {IEEE} {ICIP} 2014, Paris, France, October 27-30, 2014},
  2014, pp. 2709--2713.

\bibitem{DBLP:conf/icml/DuchiSSC08}
J.~C. Duchi, S.~Shalev{-}Shwartz, Y.~Singer, and T.~Chandra,
\newblock ``Efficient projections onto the \emph{l}\({}_{\mbox{1}}\)-ball for
  learning in high dimensions,''
\newblock in {\em Proceedings of {ICML} 2008}, 2008, pp. 272--279.

\bibitem{DBLP:conf/icassp/AfrasiyabiBNYYC18}
A.~Afrasiyabi, D.~Badawi, B.~Nasir, O.~Yildiz, F.~T. Yarman{-}Vural, and
  A.~Enis {\c{C}}etin,
\newblock ``Non-euclidean vector product for neural networks,''
\newblock in {\em 2018 {IEEE} {ICASSP} 2018}, 2018, pp. 6862--6866.

\bibitem{DBLP:journals/tsp/KopsinisST11}
Y.~Kopsinis, K.~Slavakis, and S.~Theodoridis,
\newblock ``Online sparse system identification and signal reconstruction using
  projections onto weighted ell\({}_{\mbox{1}}\) balls,''
\newblock {\em {IEEE} Trans. Signal Processing}, vol. 59, no. 3, pp. 936--952,
  2011.

\bibitem{10.2307/2291512}
D.~L. Donoho and I.~M. Johnstone,
\newblock ``Adapting to unknown smoothness via wavelet shrinkage,''
\newblock {\em Journal of the American Statistical Association}, vol. 90, no.
  432, pp. 1200--1224, 1995.

\bibitem{DBLP:journals/corr/RadfordMC15}
A.~Radford, L.~Metz, and S.~Chintala,
\newblock ``Unsupervised representation learning with deep convolutional
  generative adversarial networks,''
\newblock in {\em {ICLR} 2016}, 2016.

\bibitem{DBLP:journals/corr/MolchanovTKAK16}
P.~Molchanov, S.~Tyree, T.~Karras, T.~Aila, and J.~Kautz,
\newblock ``Pruning convolutional neural networks for resource efficient
  transfer learning,''
\newblock {\em CoRR}, vol. abs/1611.06440, 2016.

\bibitem{DBLP:journals/corr/abs-1710-09282}
Yu~Cheng, Duo Wang, Pan Zhou, and Tao Zhang,
\newblock ``A survey of model compression and acceleration for deep neural
  networks,''
\newblock {\em CoRR}, vol. abs/1710.09282, 2017.

\bibitem{DBLP:conf/date/PandaSR16}
Priyadarshini Panda, Abhronil Sengupta, and Kaushik Roy,
\newblock ``Conditional deep learning for energy-efficient and enhanced pattern
  recognition,''
\newblock in {\em Design, Automation {\&} Test in Europe Conference {\&}
  Exhibition}, 2016, pp. 475--480.

\bibitem{biasielli1}
M.~Biasielli, C.~Bolchini, L.~Cassano, E.~Koyuncu, and A.~Miele,
\newblock ``A neural network based fault management scheme for reliable image
  processing,''
\newblock {\em Submitted for publication}, Apr. 2019.

\end{thebibliography}

\end{document}